\documentclass[journal]{IEEEtran}

\usepackage{ifpdf}
\usepackage{amsmath,amsfonts}
\usepackage{amssymb}  
\usepackage{algorithmic}
\usepackage{algorithm}
\usepackage{array}
\usepackage[caption=false,font=footnotesize]{subfig}
\usepackage{textcomp}
\usepackage{stfloats}
\usepackage{url}
\usepackage{verbatim}
\usepackage{graphicx}
\usepackage{cite}
\usepackage{booktabs}
\usepackage[export]{adjustbox}
\usepackage{bm}
\usepackage{nccmath}
\usepackage{mathtools}
\usepackage{hyperref}
\usepackage{tensor}
\usepackage[dvipsnames]{xcolor}
\usepackage{units}
\usepackage{tikz-cd}
\usepackage{bbm}
\usepackage{tikz}
\usepackage{cases}

\newcommand{\R}{\mathbb{R}}

\hypersetup{hidelinks}

\begin{document}

\title{Dynamic Mobile Manipulation via Whole-Body Bilateral Teleoperation of a Wheeled Humanoid}

\author{{Amartya Purushottam$^1$, Yeongtae Jung$^2$, Christopher Xu$^1$, and Joao Ramos$^{1,3}$}
\thanks{$^{1,3}$The authors are with the $^1$Department of Electrical and Computer Engineering and the $^3$Department of Mechanical Science and Engineering at the University of Illinois at Urbana-Champaign, USA.}
\thanks{$^2$The author is with the Department of Mechanical System Engineering and Advanced Transportation Machinery Research Center, Jeonbuk National University, Jeollabuk-do 54896, Republic of Korea.}
\thanks{This work is supported by the National Science Foundation via grant IIS-2024775.}}

\maketitle

\begin{abstract}
Humanoid robots have the potential to help human workers by realizing physically demanding manipulation tasks such as moving large boxes within warehouses. We define such tasks as Dynamic Mobile Manipulation (DMM).
This paper presents a framework for DMM via whole-body teleoperation, built upon three key contributions: Firstly, a teleoperation framework employing a Human Machine Interface (HMI) and a bi-wheeled humanoid, SATYRR, is proposed. Secondly, the study introduces a dynamic locomotion mapping, utilizing human-robot reduced order models, and a kinematic retargeting strategy for manipulation tasks. Additionally, the paper discusses the role of whole-body haptic feedback for wheeled humanoid control. Finally, the system's effectiveness and mappings for DMM are validated through locomanipulation experiments and heavy box pushing tasks. Here we show two forms of DMM: grasping a target moving at an average speed of 0.4 m/s, and pushing boxes weighing up to 105\% of the robot's weight. By simultaneously adjusting their pitch and using their arms, the pilot adjusts the robot pose to apply larger contact forces and move a heavy box at a constant velocity of 0.2 m/s. 


\end{abstract}

\begin{IEEEkeywords}
    Humanoid whole-body control, telerobotics and teleoperation, human and humanoid motion analysis and synthesis, human-in-the-loop, motion retargeting
\end{IEEEkeywords}

\IEEEpeerreviewmaketitle

\section{Introduction}



\IEEEPARstart{C}argo, freight, and heavy-lift accidents are common occurrences in the maritime, transportation, and warehousing industries. While manually slotting cargo containers on ships, workers are at risk of injuring their backs and being hurt by loose cargo that may be set in motion due to the rocking of the boat. With growing safety and labor concerns, humanoid robots are a potential solution that demands exploration. To be effective supportive tools, they must be capable of Dynamic Mobile Manipulation \textbf{(DMM)}:  using whole body coordination, manipulation, and locomotion to regulate and amplify forces on the environment. For example, when humans move heavy boxes, they push with their arms and use their weight by leaning with their body. Here, DMM encapsulates the human strategy of adjusting their body and arm pose, using their body weight to maximize applied effort, and regulating contact forces and acceleration of the object. These types of capabilities have yet to be seen on robotic platforms even after decades of incredible progress in perception, planning, and control within the robotics community. Human intuition and sensorimotor skills remain unmatched. 
Teleoperation presents a promising opportunity for bridging this gap by transferring human control to remote robots. In parallel autonomous pursuits, we must also attempt to leverage human intelligence and motor control via teleoperation as a step towards DMM. 

\begin{figure}[t]
    \centering
    \includegraphics[width = \columnwidth]{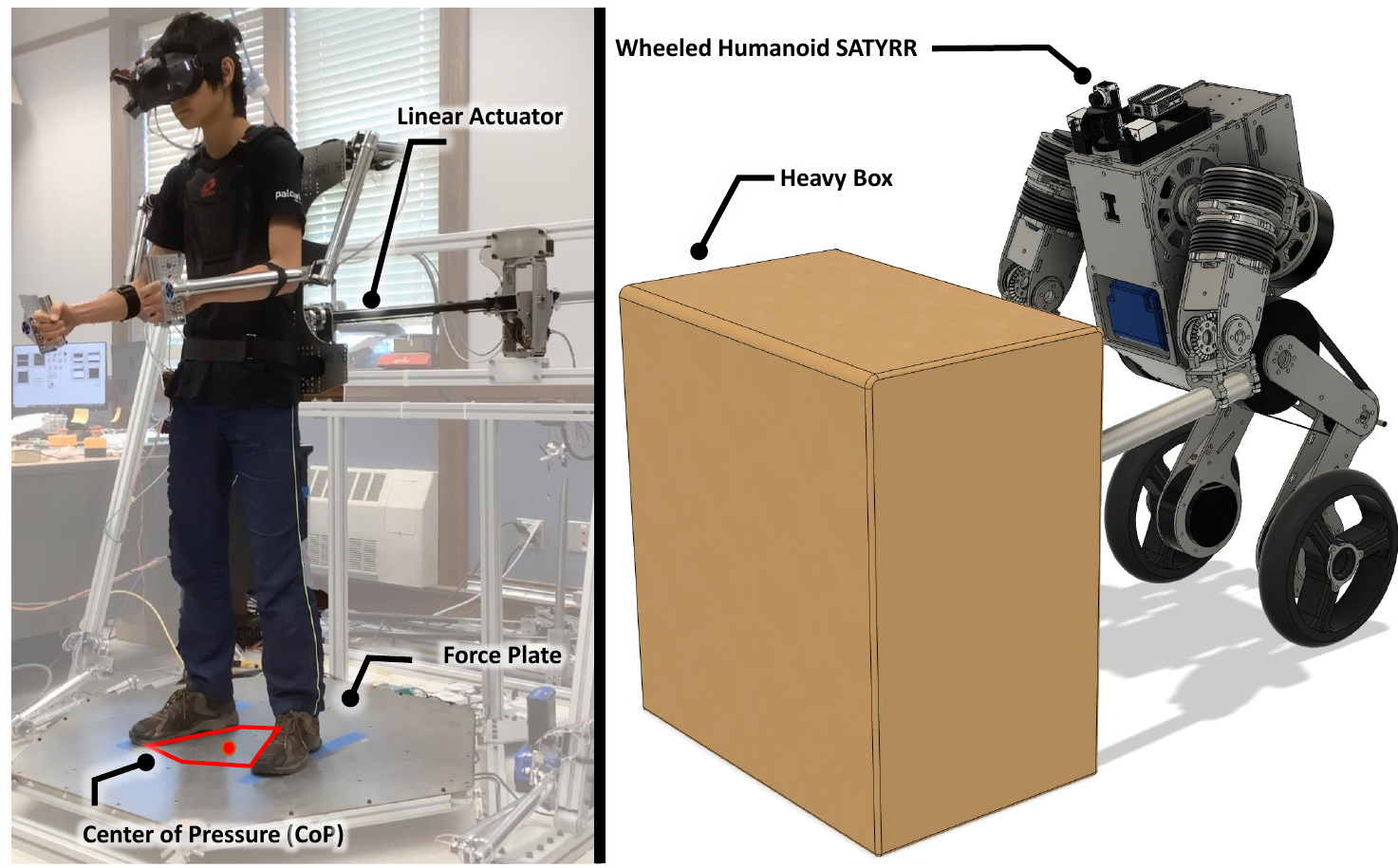}
    \caption{Dynamic Mobile Manipulation: The pilot inside the HMI (on left) teleoperates SATYRR (on right) to push a heavy box of unknown mass. Linear actuators deliver haptic feedback to human torso informing them of the robot-environment interacction forces. Demo video: https://youtu.be/QqcJsSH0YjY }
    \label{fig:MoneyShot}
    \vspace{-1.25em}
\end{figure}


We address this problem and show a form of DMM through heavy-box pushing tasks that require robot whole-body coordination to amplify and regulate applied forces. Here, the robot is unable to push the box without synergistic use of its body and arms. One approach to transporting this level of control to teleoperated humanoids is to create an immersive experience where the robot behaves as an extension of the human via motion similarity and retargeting strategies \cite{WB_Motion_Retarget,dallard2023synchronized}. This way we can utilize human body motion to generate poses and trajectories that the robot can track and follow. More specifically, to enable hands-free \cite{Puru2022}, dynamic manipulation and locomotion control we envision that the operator will use their arms for robot manipulation while simultaneously controlling robot locomotion by adjusting their body lean. To achieve this, we capture the human Center of Mass (CoM) and arm motion through an exoskeleton frame and Human Machine Interface (HMI) \cite{wang2021dynamic}. To further the immersive experience and provide the pilot information about the robot's surroundings we deliver haptic feedback to the human CoM to allow the human to feel similar forces to those experienced by the robot. This bilateral teleoperation framework has been shown to be effective in transferring human motion control for humanoid telelocomotion \cite{ramos2018humanoid,tablisbilateral}. \\


The primary contributions of this study are:
\begin{itemize}
    \item Development of a model-based dynamic locomotion mapping using human-robot Reduced Order Models (RoM) and a kinematic manipulation retargeting strategy
    \item Formulation of a haptic feedback law that gives the pilot increased pitching range of motion and enables them to feel robot external forces during manipulation for an improved immersive experience
    \item Experimental validation of our whole-body teleoperation framework for DMM in the context of heavy (105\% robot weight and unknown to pilot) box pushing.
\end{itemize}


\section{Related work} \label{sec:related_work}
 Researchers have addressed the locomotion and manipulation of humanoids by developing controllers using full-body dynamics \cite{ali_humanoid,dynamicwheeledhumanoid}, or reduced-order models (RoM) such as the Wheeled Inverted Pendulum (WIP) \cite{wensing2020wheeled,wheeledhumanoid2019model}, and Linear Inverted Pendulum (LIP) \cite{colin2022bipedal}. However, these works did not demonstrate interaction with the robot's surroundings -  a prerequisite of DMM. Minniti \emph{et al.} in \cite{huttergreatwork} showcased impressive dexterity and interaction in performing mobile manipulation tasks with their balbot manipulator, using model predictive control and an adaptive parameter estimator to open doors and lift objects. Stillman \emph{et al.} presented the Golem Krang \cite{stilman2010golem}, a bi-wheeled manipulator capable of balancing while performing heavy lifting. Other related works \cite{hutter_vision} have integrated vision pipelines to facilitate mobility and manipulation. While these studies show promising preliminary forms of DMM, they do do not leverage human adaptability, and are constrained in their capacity to modify the robot's posture to synergistically combine the robot's body and arms for force amplification. Recently, researchers proposed a framework that generates poses offline for heavy box pushing experiments, utilizing the robot's inertia and arm manipulation for DMM \cite{li2023kinodynamics}. The authors here rely on significant assumptions about knowledge of the environment (object mass, inertia, global pose, etc.) and require offline pose optimization prior to execution.

Teleoperated systems offer a solution to the aforementioned challenge of offline trajectory and pose generation by utilizing human intuition and planning. Previous works have demonstrated mobile manipulation through whole-body teleoperation \cite{teleop_MOCA,joystickHumanoidTeleop}. These systems, however, are constrained by predefined switching modes for locomotion and manipulation or limited control over individual sections of the robot. Such motion retargeting strategies restrict the dynamic capabilities of these platforms. Although real-time whole-body motion retargeting strategies from humans to humanoids have been proposed \cite{WB_Motion_Retarget,motion_retarget_wContact}, they have only been explored in simulation or for statically stable robots.

Moreover, physical differences between humans and robots, such as size and mass, make it challenging to create direct, one-to-one mappings between human and robot movements. Researchers have explored the use of whole-body haptic feedback \cite{ProfPaper2,tablisbilateral,colin2022bipedal} to bridge this gap between human and robot dynamics. These works have yet to explicitely show the effects of interaction forces between the robot and its environment on the humans feedback force.


\section{Background}
\label{sec:background}
Retargeting of individual limbs and whole-body movements is challenging due to human-robot kinematic differences \cite{motion_retarget_wContact}, and the increased computational cost associated with controlling larger nonlinear models. Therefore, we use RoM that reduce modeling complexity while still capturing the key unstable dynamics of the robot and human. The human's arms and body motion are used to generate retargeted trajectories, eliminating the need for a joystick in this hands-free teleoperation framework \cite{Puru2022}. Using body pitch, as shown in Fig. \ref{fig:Human-Robot Actuated Pendulum}, the pilot can directly control interaction forces between the robot and its environment. Force control is particularly beneficial within environments with unknown dynamics (e.g. box mass, inertia, friction, etc) \cite{lynch2017modern}.

Here, we briefly revisit the robot cart-pole RoM used for mappings, and the human DCM that captures their dynamic balancing strategy.

\subsection{Cart-Pole Model}
The dynamics of the cart-pole system, linearized around the upright zero position $(sin(\theta_R) \approx \theta_R ; \: cos(\theta_R) \approx 1)$, with small pendulum velocities $(\dot{\theta}^2_R \approx 0)$ can be written in state-space form \cite{cartpole}:

\begin{equation} 
    \label{linStateSpaceModel}
    \begin{bmatrix}
        \dot{x}_w\\ \ddot{x}_w\\ \dot{\theta}_R\\ \ddot{\theta}_R
    \end{bmatrix} \!\!=\!\! 
    \begin{bmatrix}
        0\!\! & 1\!\! & 0\! & 0\\
        0\!\! & 0\!\! & -\frac{m}{M}g\! & 0\\
        0\!\! & 0\!\! & 0\! & 1\\
        0\!\! & 0\!\! & \frac{mg}{Mh_R}\! & 0
    \end{bmatrix}\!\!\!\!
    \begin{bmatrix}
        x_w\\ \dot{x}_w\\ \theta_R\\ \dot{\theta}_R
    \end{bmatrix} \!\!+\!\!
    \begin{bmatrix}
        0 \\
        \frac{1}{M}\\
        0\\
        \frac{-1}{M h_R}
    \end{bmatrix} \!\! u
    +\!
    \begin{bmatrix}
        0 \\
        0\\
        0\\
        \frac{1}{mh_R}
    \end{bmatrix}\!\!F_{ext}
\end{equation}
where $x_w$ represents the horizontal position of the cart, $\theta_R$ the angle between the cart and CoM, $h_R$ the pendulum height, $g$ the gravity constant, $u$ the force applied to the cart, and $F_{ext}$ a disturbance force applied at the CoM as seen in Fig. \ref{fig:Human-Robot Actuated Pendulum}. The mass of the cart base, $M$, for humanoid robots can assumed to significantly smaller than the the mass of the body, $m$. In other words, $m >> M$. 

Here, it can be seen that controlling desired base acceleration is proportional to controlling the robot's lean angle: $    \ddot{x}_w^{des} \propto \theta^{des}_R $.
In other words, the pilot can control translational forces exerted on the environment by the robot through regulation of the robot's pitch.


\subsection{Divergent Component of Motion (DCM)}
Human motion for balancing and locomotion can be sufficiently captured by pendular systems such as the LIP \cite{colin2022bipedal}. The pendular DCM is the unstable component of the dynamics that we assume also captures the human's balancing strategy \cite{dynStCond}. Specifically, human motion can be stabilized as long as the LIP DCM remains within the human support polygon: $p_x^{min}  \leq x_o + \frac{\dot{x}_o}{\omega_o} \leq p_x^{max}$. Here, $x_o$ is the CoM horizontal position, $\omega_o = \sqrt{g/h}$ the human pendulum's natural frequency at height $h$ under a gravity field $g$, and $p_x^{min}, p_x^{max}$ are the Center of Pressure (CoP) minimum-maximum given by the human foot length. Under a change of state, $\theta_o = x_o/h$ (valid around the linearized zero position), this condition can be rewritten: 
\begin{equation}
  \frac{p_x^{min}}{h} \leq \theta_o + \frac{\dot{\theta}_o}{\omega_o} \leq \frac{p_x^{max}}{h}
\end{equation}
We refer to the dimensionless pendular DCM as $\xi = \theta_o + \frac{\dot{\theta}_o}{\omega_o}$. For brevity, $\xi$ will simply be referred to as the DCM.

\section{Kinodynamic Whole-body Motion Retargeting} \label{sec:methods}


Here, we introduce the whole-body bilateral teleoperation mappings for DMM. The lower body telelocomotion mapping models the human and robot as pendulum systems along the sagittal plane. Section \ref{subsec:HumanModel} motivates the human fixed-based actuated inverted pendulum (AIP) model. Section \ref{subsec:RobotModel} defines a condition for stabilizing the robot cart-pole given by its dimensionless DCM. We use the human's dynamic balancing strategy given by their DCM and CoP to generate tracking references and haptic feedback as shown in Sections \ref{subsec:DynSim} and \ref{subseq:virtual_spring}. The upper body telemanipulation mapping outlined in section \ref{subsec:Arm_Mapping}, uses similarity in human-robot arm joint topology to enable intuitive kinematic arm control. The focus lies on performing DMM through the proposed mappings and feedback, and does not necessarily claim exclusivity. The pilot's adaptability to different mappings and feedback present a challenge in establishing necessity within this context.


 \vspace{-1.0em}
\subsection{Human Motion Model} \label{subsec:HumanModel}

Use of human pitch for hands-free teleoperation and control of wheeled humanoids has shown viability \cite{Puru2022}. Attempting to capture the pitch angle from classically used human locomotion models, such as the LIP, presents a problem however. This  angle - defined between the human CoP and CoM - converges to zero in steady state as the CoP tracks the CoM ($p_H \rightarrow x_H$).
This results in stop-and-go motions and unintuitive pitch control on wheeled robots. For example, the human may lean at a constant angle, but the desired reference is zero causing the robot be upright.

\begin{figure}[t]
    \centering
    \includegraphics[width = \columnwidth]{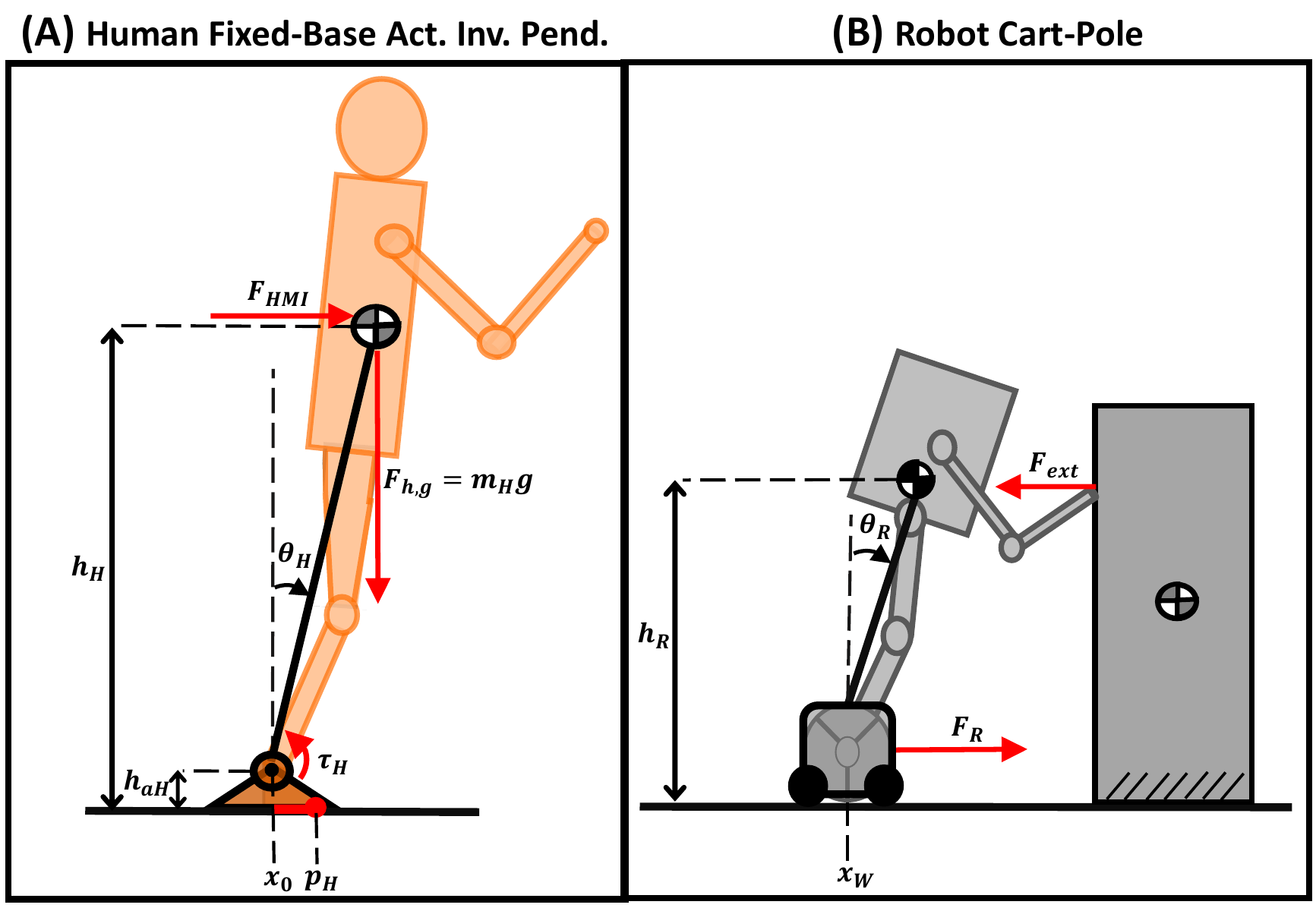}
    \caption{\textbf{(A)} The human is modeled as an actuated pendulum where the ankle as treated as an actuator to control lean. \textbf{(B)} The robot is modeled as cart-pole for generating desired references}
    \label{fig:Human-Robot Actuated Pendulum}
    \vspace{-1.25em}
\end{figure}

Thus, we propose applying a transformation to model the human as a fixed-base actuated inverted pendulum (\textbf{AIP}) as shown in Fig. \ref{fig:Human-Robot Actuated Pendulum}. This in turn allows us to define the pitch angle between the fixed human ankle and their CoM. This measurement remains constant in steady state and does not change as a function of the CoP. Instead, the CoP is used to define the ankle torque needed to stabilize the human inverted pendulum. 


The dynamics for the AIP are given by:
\begin{equation}
   \Ddot{\theta}_{H}=\frac{g \sin(\theta_{H})}{\tilde{h}_H }+\frac{1}{m_H\tilde{h}_H^2}\tau_{H}+\frac{\cos(\theta_{H})}{m_H\tilde{h}_H}F_{HMI}
\end{equation}
where $m_H$ is the human's mass, and $\tilde{h}_H = h_H - h_{aH}$ is the difference in the CoM height, $h_H$, and ankle height, $h_{aH}$. The human pitch angle, $\theta_{H}$, and human ankle torque, $\tau_H$, are defined in the clockwise direction. The external feedback force, $F_{HMI}$, is applied to the human torso near their CoM. Human motion is measured and force feedback is applied in realtime  through this HMI \cite{wang2021dynamic,Puru2022}.

Linearizing around the $\theta_H=0$ upright position, and assuming $h_{H}\gg h_{aH}$ allows us to write in state-space form:
\begin{equation}
    \label{eq:human_aip_dyn}
    \begin{bmatrix}
        \dot{\theta}_{H}\\ \ddot{\theta}_{H}
    \end{bmatrix} = 
    \begin{bmatrix}
        0 & 1\\
        \omega_{H}^2 & 0
    \end{bmatrix}\!\!\!
    \begin{bmatrix}
        \theta_H\\ \dot{\theta}_H
    \end{bmatrix} +
    \begin{bmatrix}
        0 \\
        \frac{1}{m_Hh^{2}_{H}}
    \end{bmatrix} \tau_{H}
     +
    \begin{bmatrix}
        0 \\
        \frac{1}{m_Hh_{H}}
    \end{bmatrix} F_{HMI}
\end{equation}
The state transition matrix has eigenvalues $\pm \omega_H$. Diagonalizing as shown in \cite{caroneDcM} and substituting human actuation as a function of their CoP ($\tau_H = p_{H}m_H g$) we find the unitless divergent component of motions and its time derivative:


\begin{align}
    \xi_H &= \theta_H + \tfrac{1}{\omega_H}\dot{\theta}_H 
    \\
    \dot{\xi}_H &=\omega_H \xi_H - \tfrac{\omega_H}{h_H}p_H
    \label{eq:human_dcm_and_dynamics}
\end{align}

\vspace{-1.25em}
\subsection{Robot Cart-Pole DCM} \label{subsec:RobotModel}
To track the human pendular DCM we isolate the linearized cart-pole DCM here. The eigenvalues of Eq. \ref{linStateSpaceModel} are:
$(\lambda_{x_w}, \lambda_{\dot{x}_w}, \lambda_{\theta_R}, \lambda_{\dot{\theta}_R}) = (0, 0, \sqrt{g\slash h_R}, -\sqrt{g\slash h_R} )$.
The states corresponding to eigenvalues equal to 0 indicate marginally stable behavior in these modes. In other words, these states neither grow nor decay when perturbed. To design control references that can stabilize the unstable dynamics we focus on isolating the non-zero eigenvalue variables $\theta_R$, and $\dot{\theta}_R$:
\begin{equation} \label{dyn_of_int}
    \begin{bmatrix}
        \dot{\theta}_R\\ \ddot{\theta}_R
    \end{bmatrix} = 
    \begin{bmatrix}
        0 & 1\\
        \alpha^2\omega_R^2 & 0
    \end{bmatrix}
    \begin{bmatrix}
        \theta_R\\ \dot{\theta}_R
    \end{bmatrix} +
    \begin{bmatrix}
        0 \\
        \frac{-1}{Mh_R}
    \end{bmatrix} F_R
    +
    \begin{bmatrix}
        0 \\
        \frac{1}{mh_R}
    \end{bmatrix} F_{ext}
\end{equation}
where $\omega_R = \sqrt{g/h_R}$ is the robot linear natural frequency and $\alpha = \sqrt{m/M}$ is a unitless quantity from Eq.~(\ref{linStateSpaceModel}).

Performing the same diagonalization and decomposition as in the previous section, letting $\omega_R^{\circ} = \alpha\omega_R$ (since $\alpha$ is unitless), we find the cart-pole DCM and its dynamics:
\begin{align}
    \label{eq: Robot_dcm_and_dynamics}
         \xi_R &=\theta_R + \tfrac{1}{\omega_R^{\circ}}\dot{\theta}_R
         \\ 
         \dot{\xi}_R &=
         \omega_R^{\circ} \xi_R -\tfrac{1}{\omega_R^{\circ} Mh_R}F_R
\end{align}

 

\subsection{Dynamic Locomotion Retargeting}\label{subsec:DynSim}

\begin{figure*}[t] 
    \centerline{\includegraphics[width=17.45cm]{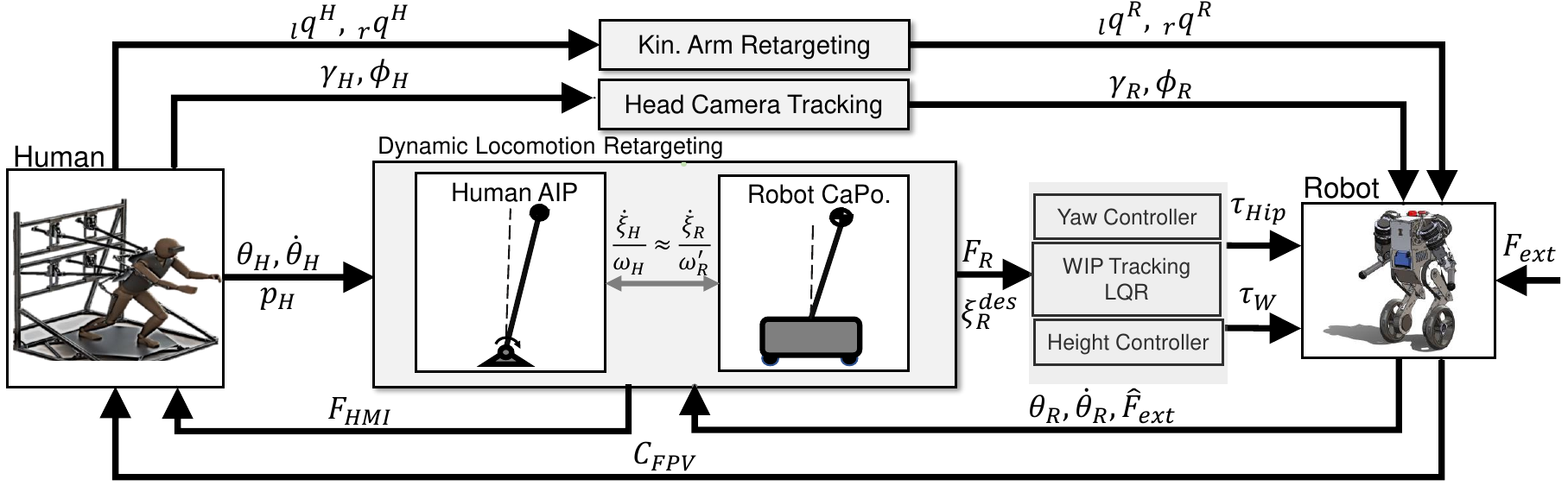}}
    \caption{(A) Full system layout and bilateral teleoperation control framework. The real-time first-person view camera feedback is shown as $C_{FPV}$. The prescripts $l,r$ represent the left and right hands. $\gamma_H$ and $\phi_H$ represent the human head pitch and yaw.}
    \label{fig:HMI_wholesys}
    \vspace{-1.25em}
\end{figure*}

The proposed tele-locomotion mapping accounts for the different sizes of the robot and human by normalizing along their natural frequency, masses, and heights \cite{ProfPaper2}. The tracking of the DCM is a design choice in this framework motivated by its ability to capture the dynamic balancing requirement of the human pendulum model. By enforcing the robot follow the human's DCM and its time derivative we can:
\begin{itemize}
\item Stabilize and dynamically control the robot using the AIP DCM as a tracking reference
\item Capture the non-minimum phase behavior of the human CoP as a feedforward for increased responsiveness  
\item Explicitly model and map virtual and external forces acting on the human and robot
\end{itemize}
We choose the DCM as our retargeted state and aim to achieve: 
\begin{equation} \label{dcm_kin_sim}
    \begin{gathered}[b]
        \xi_R = \xi_H  \\
    \end{gathered}
    \vspace{-0.5em}
\end{equation}
To minimize the DCM error, $|\xi_R - \xi_H|$, while accounting for the different human-robot sizes, we try to keep the normalized time derivative \cite{ProfPaper2} of the DCMs identical: 
\begin{equation} \label{Dyn_constraint}
\frac{\dot{\xi}_R}{\omega^\circ_R} = \frac{\dot{\xi}_H}{\omega_H}
\end{equation}
where $\dot{\xi}_R = \dot{\theta}_R + \frac{\ddot{\theta}_R}{\omega_R^{\circ}}$ and $\dot{\xi}_H = \dot{\theta}_H + \frac{\ddot{\theta}_H}{\omega_H}$ are the time derivatives of the DCM in Eq.~\eqref{eq:human_dcm_and_dynamics} and Eq.~\eqref{eq: Robot_dcm_and_dynamics}. This constraint Eq.~\ref{Dyn_constraint} couples the angular dynamics of the human and robot as seen by expanding and substituting in Eq.~(\ref{eq:human_aip_dyn}) and Eq.~(\ref{dyn_of_int}):

\begin{equation} \label{eq:exp_dyn_sim}
        \theta_R \!+\! \frac{\dot{\theta}_R } {\omega_R^{\circ}} \!-\!\frac{F_R}{\gamma_R} \!+\! \frac{F_{ext}}{\alpha^{2} \gamma_R} = 
        \theta_H \!+\! \frac{\dot{\theta}_H }{\omega_H} \!-\! \frac{p_H}{h_H} \!+\! \frac{F_{HMI}}{\gamma_H} 
\end{equation}
where $\gamma_{j} = m_{j}\omega_{j}^2 h_{j}$ with $j = \{H,R\}$ has units Newtons and can be viewed as a nondimensionalizing scaling for human-robot forces. Although there are many choices of $F_{HMI}$ and $F_R$ that satisfy Eq.~(\ref{eq:exp_dyn_sim}), we propose a specific formulation where the human's applied effort, captured by their CoP, is analogous to the force or effort applied on the robot cart-pole:  
\begin{equation}\label{eq:dyn_fb_a}
    F_R = \frac{\gamma_R}{h_H} p_H
\end{equation}
The remaining terms in Eq.~(\ref{eq:exp_dyn_sim}) are fed back to the human to provide information about the robots states and interaction forces for an improved immersive experience:
\begin{equation}\label{eq:dyn_fb_b}
F_{HMI} \!=\! \gamma_H \! \left(\!\!(\theta_R \!-\! \theta_H) \!+\! \!\left(\frac{\dot{\theta}_R}{\omega^{\circ}_R} \!-\! \frac{\dot{\theta}_H}{\omega_H}\!\!\right)\!\!\right) \!+\! \frac{\gamma_H}{\alpha^2 \gamma_R}F_{ext}     
\end{equation}
More specifically, the feedback force tries to minimize the difference in the human-robot angular motions and also conveys the robot's external forces scaled by human-robot parameters. The robot force, $F_R$, captures the non-minimum phase behavior of the CoP and is particularly valuable for bi-wheeled humanoids as their base must also initially move in the opposite direction compared to the desired reference.  

A WIP Linear Quadratic Regulator (LQR), with controller states $[x_W, \dot{x}_W, \xi_R]^T$, is used to track the human DCM as shown in Fig. \ref{fig:HMI_wholesys}. To transfer the human's intended motion, improve responsitivity, and satisfy Eq.~(\ref{eq:exp_dyn_sim}) (under perfect tracking), $F_R$ is treated as a feedforward force. The resulting wheel control effort is given by:
\begin{equation}
     \tau_W = -K_{LQR}(\xi_H - \xi_R) + F_R
\end{equation}
 Tracking of wheel position and velocity is intentionally disabled here as it can result in accumulation of error and wheel slip when interacting with unknown objects and forces e.g. pushing heavy boxes of unknown mass. Therefore, these states are intentionally not tracked. Moreover, solely regulating the DCM allows the pilot to control interaction forces with the environment.

\vspace{-1.0em}
\subsection{Virtual Spring Force} \label{subseq:virtual_spring}
To allow the pilot to dynamically lean further without falling over, a safety force was incorporated as a virtual spring, $F_s =-K_s x_H$, where $x_H$ is the human CoM displacement from their upright position. This is modeled by adding $F_{s}/\gamma_H$ to both sides of Eq.~(\ref{eq:exp_dyn_sim}).
Here, $K_{s} = 400$ N/m is tuned by pilot preference. The resulting net feedback and feedforward and feedback terms are: 

\begin{subequations}
\begin{equation}
\label{eq:net_feedback_a}
    F_R \leftarrow F_R - \frac{\gamma_R}{\gamma_H}F_{s}
\end{equation}
\begin{equation}
\label{eq:net_feedback_b}
    F_{HMI} \leftarrow F_{HMI} + F_{s}
\end{equation}
\end{subequations}

The resulting haptic force, $F_{HMI}$, includes a virtual spring. Consequently, when the human leans forward, $F_{s} < 0$, the robot feedforward, $F_R$ is also larger. This term encapsulates how much effort the human exerts to resist the spring. For box pushing, after the robot's pose (it's lean and contact locations) is chosen by the pilot, this augmented feedforward allows the robot to apply more effort via its wheels.


\subsection{Kinematic Manipulation Retargeting} \label{subsec:Arm_Mapping}
 We leverage similarity in joint topology of the human and robot arms to develop a intuitive kinematic mapping. We approximate robot and human shoulders as spherical joints to simplify the inverse kinematics (IK) problem.

\begin{figure}[t]
    \centering
    \includegraphics[width = \columnwidth]{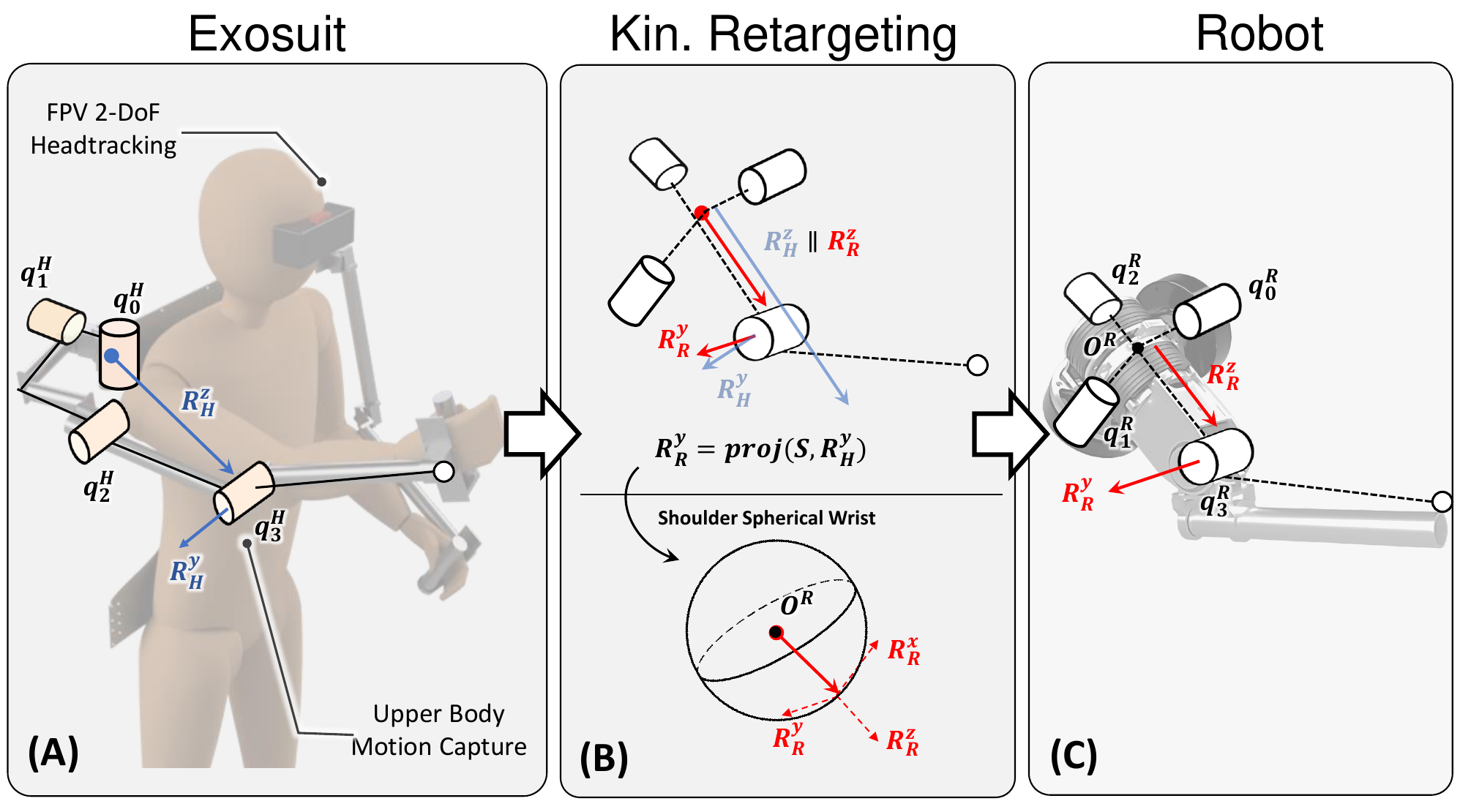}
    \caption{Human arm motion is captured by an exoskeleton suit. A design choice was made to align the human-robot upper arm and elbow rotation axis in the shoulder spherical wrist mapping. }
    \label{fig:arm_retargeting}
    \vspace{-1.25em}
\end{figure}

We define the human and robot arm generalized  coordinates:
\begin{equation}
    \bm{q}^H \!=\! [q_0^H \; q_1^H \; q_2^H \; q_3^H]^T ,\;
    \bm{q}^R \!=\! [q_0^R \; q_1^R \; q_2^R \; q_3^R]^T 
\end{equation}
where $q_0$ to $q_3$ are the measured human-robot joint angles shown in Fig. \ref{fig:arm_retargeting}. The following computations are performed for both arms, but for shorthand shown only for a single arm. The unit vector from human shoulder to elbow, $\bm{\mathcal{R}^}z_{H}$, in Fig. \ref{fig:arm_retargeting} (A) can be found using the motion capture suit's forward kinematics.

The human and robot elbows' move approximately on spheres centered around their respective shoulders. Therefore, in Fig. \ref{fig:arm_retargeting} (B), we align the unit vector from the robot's shoulder to its elbow with the direction of the human's elbow:
\begin{equation}
    \bm{\mathcal{R}}^z_{R} = \bm{\mathcal{R}}^z_{H}
\end{equation}
To find the remaining axis we define the set of vectors, $\bm{\mathcal{S}}$, perpendicular to $\bm{\mathcal{R}}^z_H$:
\begin{equation}
    \bm{\mathcal{S}} = \{\bm{\mathcal{R}}_u \;|\; {\bm{\mathcal{R}}_u} \cdot {\bm{\mathcal{R}}^z_H} = 0\} 
\end{equation}
The second axis of rotation can be subsequently set:

\begin{equation}
    \bm{\mathcal{R}}^y_R = \text{proj}_{\bm{\mathcal{S}}}(\bm{\mathcal{R}}^y_H)
\end{equation}
The aim of this projection is to find the vector closest to that of the human elbow rotation axis $\bm{\mathcal{R}}^y_H$ that is also perpendicular to the given $\bm{\mathcal{R}}^z_H$ as shown in Fig. \ref{fig:arm_retargeting} (B). The final axis is given by their cross product: $\bm{\mathcal{R}}^x_R = \bm{\mathcal{R}}^y_R \times \bm{\mathcal{R}}^z_R$. With all 3 axes of the spherical wrist defined,  we can solve the inverse kinematics of the spherical wrist as given in \cite{2021SphericalWrist}:
\begin{equation}
         \bm{q}^R_{[0:2]} = \mathrm{IK}(\bm{\mathcal{R}}^x_R, \bm{\mathcal{R}}^y_R, \bm{\mathcal{R}}^z_R)        
\end{equation}
 Similarity in human-robot elbow topology leads to direct use of human elbow angle as reference for the robot: $q^R_3 = q^H_3$.

The desired joint positions are directly sent to the motors where the driver runs an onboard tracking PD controller at 40 MHz:
\begin{equation*}
    \tau_{com} = K_p (q - q_{des}) + K_d (\dot{q} - \dot{q}_{des}) + \tau_{ff}
\end{equation*}
where $\tau_{ff} = 0$ and desired joint velocity are set to zero, $\dot{q}_{des} = 0$. No gravity compensation is added as the design of the upper-body considers proximal actuation design principles \cite{YsimTellow}. This allows us to assume the arms are nearly massless and contributes to smoothness in arm tracking.


\section{Experimental Setup} \label{sec:exp_setup}
We conducted two experiments to showcase efficacy of the proposed mappings in whole-body teleoperation and DMM. The first set of experiments evaluate tele-locomanipulation tasks including standing in place, accurately tracking arm movements, and performing a dynamic object hand-off. This last task is the primary demonstration of whole-body teleoperation in this experiment where the pilot simultaneously controls the robot's body and arms to grab a moving object. Yaw control can be used in conjunction as done in \cite{Puru2022} to enable 2D cartesian movement of the robot. Task 2 demonstrates a form of DMM through box pushing experiments. To highlight the effectiveness of the haptic feedback, $F_{HMI}$, this experiment is constrained to the sagittal plane using railings on the ground to prevent rotation of the box. 
\vspace{-1.0em}

\begin{table}[htbp]
  \centering
  \caption{Human-Robot Reduced Order Model Parameters}
  \label{table:sys_params}
  \begin{tabular}{cccccc}
    \toprule
    Parameter & Symbol & Human & SATYRR & Units & Ratio \\
    \midrule
     CoM Masses   & $m$    &  52.0 &   12.6 &  kg   & 4.13 \\
    CoM Height& $h$    &  1.10  &  0.37  & m    & 2.97 \\
Natural Frequency & $\omega$ & 2.99 & 5.15 & $s^{-1}$ & 0.58 \\
     Upper Arm Length & $l_{A}$ & 0 & 0 & m & 0\\
     Forearm Length & $l_{F}$ & 0 & 0 & m & 0\\
      Base Masses  & $M$    &  |   &   1.61 & kg & | \\ 
    \bottomrule
  \end{tabular}
\end{table}


Assuming massless arms as discussed in section \ref{subsec:Arm_Mapping}, we sum the estimated external forces on each hand:
\begin{equation}
    \hat{F}_{ext} = \bm{\Lambda}\bm{J}_{c,r}^{\dagger} (\bm{{G}} \bm{\tau}_{m,r}) + \bm{\Lambda}\bm{J}_{c,l}^{\dagger} (-\bm{{G}} \bm{\tau}_{m,l})
\end{equation}
where $r$ and $l$ subscripts denote the right and left arm measured motor torque $\bm{\tau}_{m,(\cdot)} \in \R^4$, and Moore-Penrose inverse of the contact Jacobian, $\bm{J}_{c,(\cdot)}$. The motor-space to joint-space mapping matrix is given by $\bm{{G}} = \textit{blkdiag}(-1, 1, 1, -2)$.
 The selection matrix, $\bm{\Lambda}$, isolates the $x$ component of the contact force for the planar models considered in this work.

 From our experiments, we found that scaling of the external force helped the pilot better perceive interaction forces and was their preferred choice. When leaning against a box the robot is quasi-static and its pendular accelerations $\ddot{\theta}_R$ and velocities $\dot{\theta}_R$ can be approximated to be zero. Furthermore, under perfect tracking, Eq.~(\ref{eq:dyn_fb_b}) and Eq.~(\ref{eq:net_feedback_b}) show that feedback to the human is dominated by the spring and external force. In this situation, to enable an immersive experience the haptic feedback should ideally convey the interaction force between the robot and the environment. Since the human is unable to perceive smaller changes in forces applied to their body a scaled value of the estimated external force was used:   

\begin{equation}
   \hat{F}_{ext} \leftarrow K_{fb} \hat{F}_{ext}
    \vspace{-0.5em}
\end{equation}
In this quasi-static case of pushing a heavy box, the pilot's desire for perceiving a scaled interaction force preceded their desire for generating a dynamically consistent reference for the robot. Pilot tuning set $K_{fb} = 2.5$ for all experiments.
 
\section{Results \& Discussion} \label{sec:results}
Section \ref{Exp:Grab_and_go} discusses the first set of experiments that show whole-body teleoperation under the proposed mappings. Section \ref{Exp:heavy_box_push} shows the human-robots ability in pushing heavy boxes as one form of DMM. The experiments are conducted in compliance with the requirements from the UIUC Internal Review Board (IRB). Validation shown on video\footnote{https://youtu.be/QqcJsSH0YjY}. 

\begin{figure}[t] 
    \label{fig:Grab_and_go_CoP}
    \centering
    \includegraphics[width = \columnwidth]{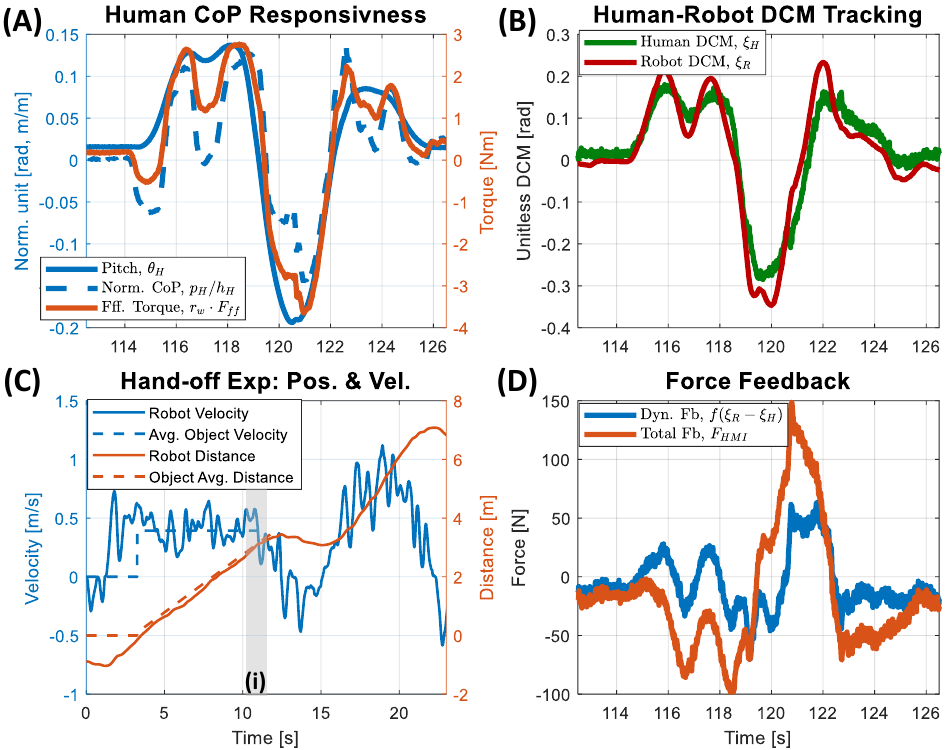}
    \caption{Whole-body tele-locomanipulation results. \textbf{(A)} highlights the non-minimum phase behavior of the CoP. \textbf{(B)} shows the robot and human DCM tracking. \textbf{(C)} shows the the dynamic hand off experiment where the robot starts at -~0.75~m, and the box starts at -~0.5~m. At \textbf{(i)} the robot catches up to the desired box position while traveling near the box's velocity. \textbf{(D)} shows the dynamic feedback and net haptic feedback as the pilot leans.}
    \vspace{-0.5em}
\end{figure}
\vspace{-1.0em}
\subsection{Whole-body Tele-locomanipulation Evaluation}
\label{Exp:Grab_and_go}

\begin{figure}[t] 
    \label{fig:Arm_Traj}
    \centering
    \includegraphics[width = \columnwidth]{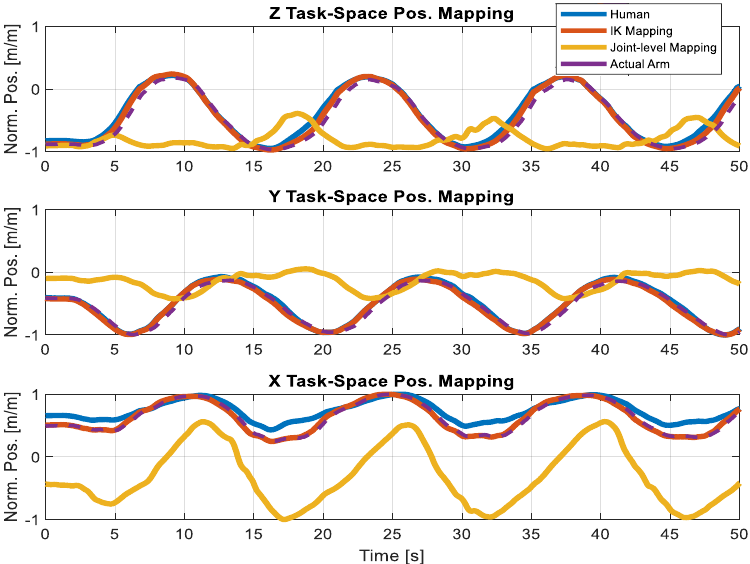}
    \caption{Arm end effector mappings and tracking: The pilot moves their arm in a circular motion in front of their body. IK tracking was preferred.}
    \vspace{-1.25em}
\end{figure}

\begin{figure}[h] \label{fig:pose_opt}
    \centering    
    \includegraphics[width=\columnwidth]{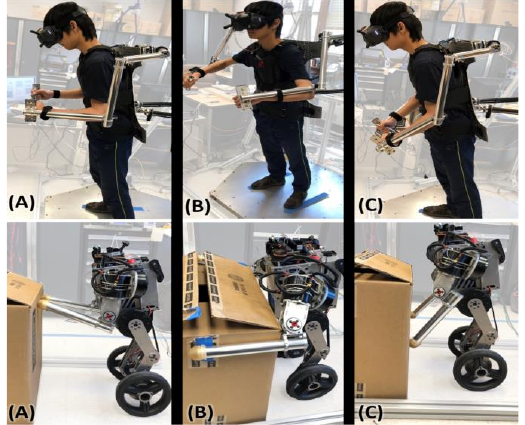}
    \caption{The pilot tried different poses during box pushing experiments and adapted their strategy to move boxes weighing 8-13kg (63.5-105\% of $m_R$). Poses \textbf{(A)}-\textbf{(B)} the pilot was unable to push the box, but \textbf{(C)} was successful.}
    \vspace{-1.75em}
\end{figure}

While standing in place or moving, locomotion control benefits from using the human CoP as a feedforward as derived in subsection \ref{subsec:DynSim}. At $t= 115$~s in Fig. \ref{fig:Grab_and_go_CoP} (A), we see how the CoP captures the non-minimum phase dynamics of human leaning (humans push of their heels to lean forward) and even precedes human pitch movement. This preemptive response results in increased responsivity as seen by the initial feedforward of $\tau_{ff} = -0.4$~Nm. Human's have a high bandwidth control of their CoP and can consequently use their feet, here, to quickly adjust the robots base or wheels. This captured human motion serves as an ideal feed-forward for the robot - which also exhibits non-minimum phase behavior.


The DCM tracking of the human-robot can be seen in Fig. \ref{fig:Grab_and_go_CoP} (B). Moving forward, the robot overshoots $\xi_H$ at $t = 122$~s and the pilot is momentarily pushed forward with a force of 50~N informing of them of the robot's overshoot. Although this feedback did not significantly contribute to stabilization, it did offer the pilot valuable situational awareness by indicating the directionality of external forces acting on the robot body. This represents a small but important step toward enhancing the immersive experience. Moreover, the pilot also tested two forms of arm mappings, where they preferred the inverse kinematic mapping. As seen in Fig. \ref{fig:Arm_Traj} the IK mapping resulted in more accurate tracking of the human hand. The slight topological difference between the human arm, the exosuit that captured the humans motion, and the robot's arm were the driving factors in this choice. This intuitive kinematic mapping ultimately enabled  hands-free tele-manipulation.


\begin{figure*}[t] 
    \centerline{\includegraphics[width=18.5cm]{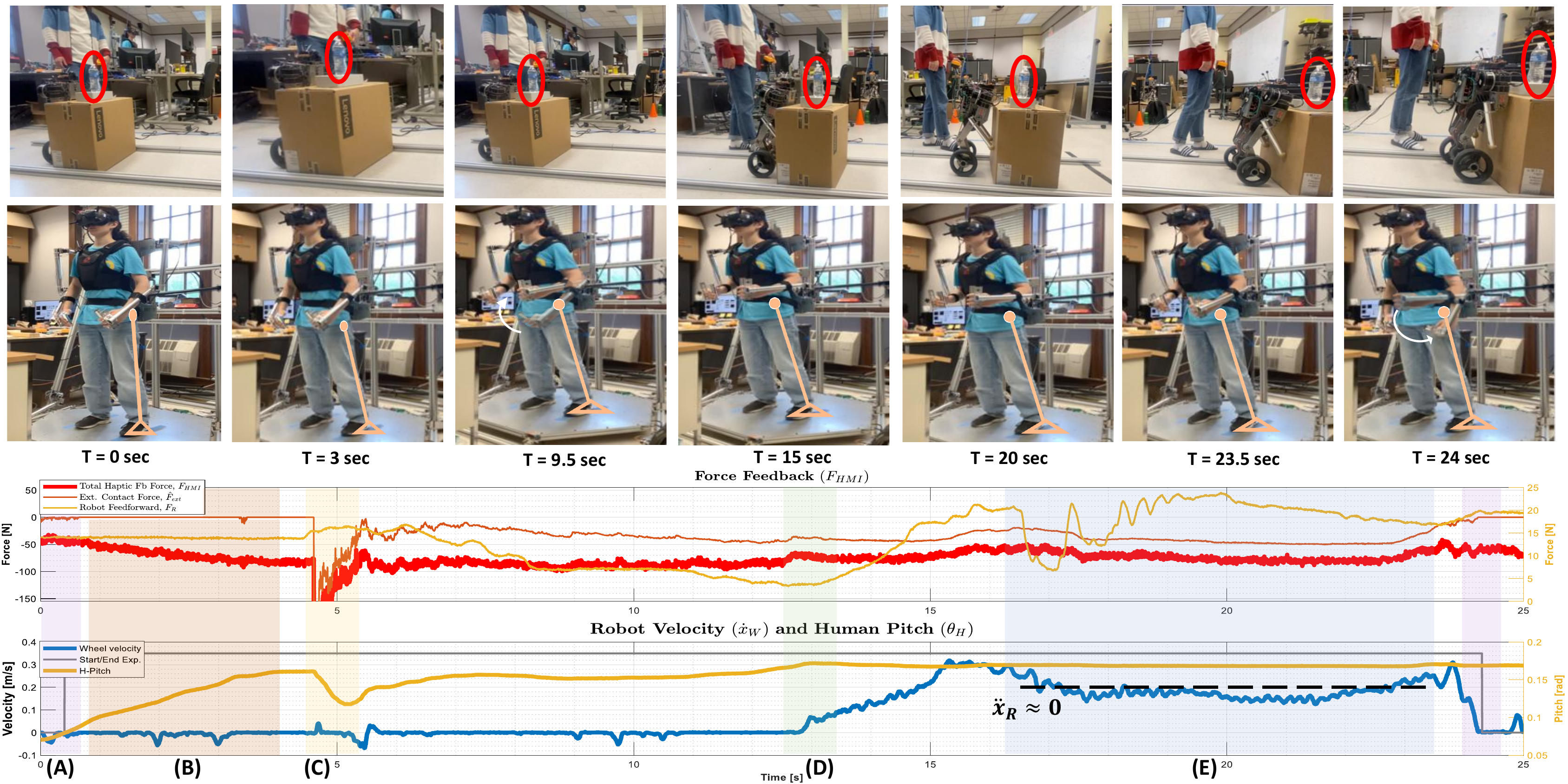}}
    \caption{Heavy box pushing experiment: A 8.5kg (67.5 \% of $m_R$) box with a balanced bottle is pushed at constant velocity. \textbf{(A)} shows in purple start and end of task. In \textbf{(B)} the pilot initially increases their lean without using their arms. At \textbf{(C)} the robot's hands make contact with the box guiding the robots hands and begins pushing with their arms resulting in a sharp change in external contact force. The pilot continues moving their arms up until \textbf{(D)}, where the robot breaks stiction and the box begins moving accompanied by a change in external force. \textbf{(E)} shows  the box moving at a constant velocity. At the end the pilot breaks contact by lowering their hands. The right axis show the zoomed in feedforward force and human pitch in their respective graphs.}
    \label{fig:BoxPushing}
    \vspace{-1.5em}
\end{figure*}

Finally, whole-body tele-locomanipulation was tested: the robot was required to grab a moving package weighing 1.25~kg (10\% of $m_R$). Completion of this task can be difficult as it requires estimation of object velocity. As shown in the grey region in Fig. \ref{fig:Grab_and_go_CoP} (C), the pilot was able to modulate the velocity to catch up to the moving package at $t = 11$~s. Without switching between locomotion and manipulation modes \cite{joystickHumanoidTeleop}, the pilot was simultaneously able to command a desired arm motion for the robot to grab the object, pull it closer to its body to mitigate the moment created around its wheel and complete the mobile hand-off. Through visual feedback the pilot could make readjustments to the grip of the object. The related video showcases the task completion. 

\vspace{-1.25em}
\subsection{Pushing heavy boxes: A form of whole-body DMM}
\label{Exp:heavy_box_push}
 The experiments here require use of whole-body coordination to regulate and apply large forces on the environment. This is shown through tasks where the robot could not succeed by leaning with their body or pushing with their arms alone. More specifically, the pilot was required to push boxes between 4-13kg at a constant velocity while keeping a bottle balanced on top of the box, as seen in Fig. \ref{fig:BoxPushing}. In this scenario, moving at a constant speed requires the pilot to regulate applied forces, while the bottle enforces bounds for the robot’s acceleration. The pilot was unaware of the box’s mass during these tests.

An initial adjustment of the robot pose was required to push heavier boxes. Finding an optimal pose for whole-body interaction and box pushing is a non-trivial task, requiring offline optimization \cite{li2023kinodynamics}. By tracking the human DCM and arm movement, the robot's stance can be adjusted on the fly as seen in Fig. \ref{fig:pose_opt}. The final chosen pose shown in Fig. \ref{fig:pose_opt} (C) maximized the achievable lean of the robot while keeping its hands closer to the bottom of the box - much like a human might. Without this modification, the robot was unable to push boxes greater than 8 kg. This adaptation highlights the effectiveness of leveraging human planning for DMM.

Fig. \ref{fig:BoxPushing} shows the pilots whole-body strategy for pushing the 8.5kg box at a constant velocity. After starting the trial run indicated by the grey line in Fig. \ref{fig:BoxPushing}, the pilot increases their lean on the box between $t = 1~s$ and $t = 4~s$. At $t = 4.5~s$ the box has not moved and the pilot telemanipulates the robot hands to begin pushing on the box. This initially results in a sharp change in contact forces as seen in Fig. \ref{fig:BoxPushing} (C). Force estimation using joint torques is particularly susceptible to noise when the velocity at the contact point is nonzero. Between $t = 5~s$ and $t = 13~s$ the pilot maintains their lean and slowly moves their arms to increase the applied force as seen by the increase in $|\hat{F}_{ext}|$. This upward arm motion also momentarily lifts the front of the box. We believe this helped in breaking static friction forces between the box and ground to start the box's movement as shown in green in Fig. \ref{fig:BoxPushing} (D) . As the robot-box move at a constant velocity - $\dot{x}_R \approx 0.2$ m/s - as highlighted in the blue section of the figure, the feedback forces are nearly constant - $F_{HMI} \approx -80N$. Upon crossing the finish line, the pilot stops by putting their hands down.

In these experiments, feedback played a nuanced but pivotal role, encompassing three components: dynamic feedback, external force, and spring force. The dynamic feedback served to inform the pilot about disparities between the human and robot DCM and the directionality of external forces applied to the robot. The external feedback force facilitated extended leaning and perception of changes in interaction forces during box pushing. The pilot found the spring feedback force preferable, as it allowed for comfortable leaning with an anticipated response and linear scaling. Without haptic feedback, i.e. $F_{HMI} = 0$, the pilot was unable to lean as far and slide the box forward. In the absence of a spring, the robot feedforward and applied force, described in Eq. (\ref{eq:net_feedback_a}), were smaller. The combination of an increased lean angle and feedforward helped the pilot break stiction. Under this bilateral teleoperation framework, various feedback laws can be implemented based on Eq. (\ref{eq:exp_dyn_sim}), offering the pilot different haptic experiences. However, the spring plays a vital role in enabling DMM and the overall success of the system.

\vspace{-0.75em}
\subsection{Limitations}
Generating position-velocity trajectories becomes challenging without knowing the box weight or friction. This difficulty arises due to the reliance on estimating noisy environmental interaction forces for trajectory planning. In previous work \cite{Puru2022}, position-velocity trajectories were generated using the linearized cart-pole steady state relationship:
\begin{equation*}
    \dot{x}_w^{des}= \int \ddot{x}_w^{des} dt = \int g\theta_H(t) dt 
\end{equation*}
Integrating desired acceleration provides velocity and position states for tracking. However, this leads to error build up and ultimately wheel slippage in the presence of external forces e.g. pushing heavy items of unknown mass. As such, the telelocomotion mappings in this work require the pilot to modulate the robot lean to indirectly control the position of the robot. This can place an undesriable cognitive load on the pilot when attempting to stay in place for prolonged periods of time. This problem could be addressed by integrating hybrid/position controllers.

Through extensive testing, our pilot could interpret the force feedback but faced challenges in discerning finer details, likely due to limited feedback resolution at the human CoM. The dynamic feedback was primarily useful for binary or tertiary classification tasks. Differentiating between changes in external forces when using a 2kg box versus a 4kg box proved challenging, while distinguishing between a 2kg and a 12kg box was feasible. This presents a potential obstacle for future whole-body bilateral teleoperation, as the dynamic feedback signal at the CoM is not easily utilized for precise regulation. Overall, our experiments demonstrated that humans excel as motion planners but are outperformed by controllers when it comes to regulation tasks.


\vspace{-0.5em}
\section{Conclusion} \label{sec:conclusion}

In this paper, we presented a framework to perform dynamic mobile manipulation via whole-body bilateral teleoperation. The locomotion mapping explicitly models external forces acting on the robot and uses the human's balancing strategy as captured by the human AIP DCM to generate tracking references. An inverse kinematics arm mapping enables the pilot to manipulate simultaneously. This becomes the foundation for performing mobile manipulation. Finally, through box pushing experiments, we show how the pilot-robot can regulate applied forces on the environment using their body and arms. The force feedback to the human enables them to perform dynamic leaning motions, and perceive the interaction forces between the environment and robot.
\vspace{-0.5em}
\section{Acknowledgment}
To my colleagues Youngwoo Sim, Guillermo Colin, and Donghoon Baek for their guidance throughout this work.


\appendices
\bibliographystyle{IEEEtran}
\bibliography{Main.bib}
\end{document}